\documentclass[11pt]{article}

\usepackage[preprint]{acl}

\usepackage{times}
\usepackage{latexsym}

\usepackage[T1]{fontenc}
\usepackage[utf8]{inputenc}
\usepackage{CJKutf8}
\usepackage{adjustbox}

\usepackage{microtype}
\usepackage{inconsolata}
\usepackage{graphicx}
\usepackage{booktabs}
\usepackage{amsmath}
\usepackage{xcolor}

\usepackage{tcolorbox}
\tcbuselibrary{breakable}

\newenvironment{statement}[1][]{%
    \begin{tcolorbox}[
        colback=blue!5,
        colframe=white,
        fonttitle=\bfseries,
        breakable,
        title={#1}
    ]%
}{%
    \end{tcolorbox}%
    \vspace{2pt}%
}

\newcommand{\JPN}[1]{\begin{CJK}{UTF8}{ipxm}#1\end{CJK}}

\title{CAT-Translate: Building Compact Open-Source Models for \\
Japanese-English Translation}

\author{Yuu Jinnai \\
  CyberAgent / Tokyo, Japan \\
  \texttt{jinnai\_yu@cyberagent.co.jp}\\
  \\
  \textbf{Models}: \url{https://huggingface.co/collections/cyberagent/cat-translate} \\
  \textbf{Dataset}: \url{https://huggingface.co/datasets/cyberagent/CAT-Translate-Dataset}
  }

\begin{document}
\maketitle

\begin{abstract}
Nowadays, large multilingual translation models demonstrate impressive translation capabilities in the machine translation benchmarks (e.g., WMT).
% Not only they achieve competitive performance on a single language pair, but they also perform well across multiple language pairs, which is a significant advantage for practitioners who need to support various languages.
This raises a practical question to the developers: is it worth developing translation models specialized for a particular language pair if you only need to support that language pair?
To give an anecdotal answer to this question, we develop a family of small language models (0.8B, 1.4B, 3.3B, and 7B parameters) specialized for Japanese-English bidirectional translation.
We employ a two-stage supervised fine-tuning approach followed by Multi-Objective GRPO~\citep{ichihara2025mo} to train models on synthetically generated parallel corpora.
% All the processes involve design choices to reduce computational requirements while maximizing translation quality, in which we find the knowledge of the source and target languages to be valuable.
We evaluate our models on WMT and real-world translation benchmarks across business, legal, medical, financial, and patent domains.
While multilingual models achieve strong performance on WMT benchmarks, our compact models outperform them on real-world benchmarks, suggesting the practical utility of developing specialized translation models even in the era of large multilingual models.
\end{abstract}

\section{Introduction}

\begin{statement}
A company needs to translate legal contracts or medical records between Japanese and English at scale, under privacy constraints, with limited GPU budget. What are their options?    
\end{statement}

\begin{figure}
    \centering
    \includegraphics[width=0.6\columnwidth]{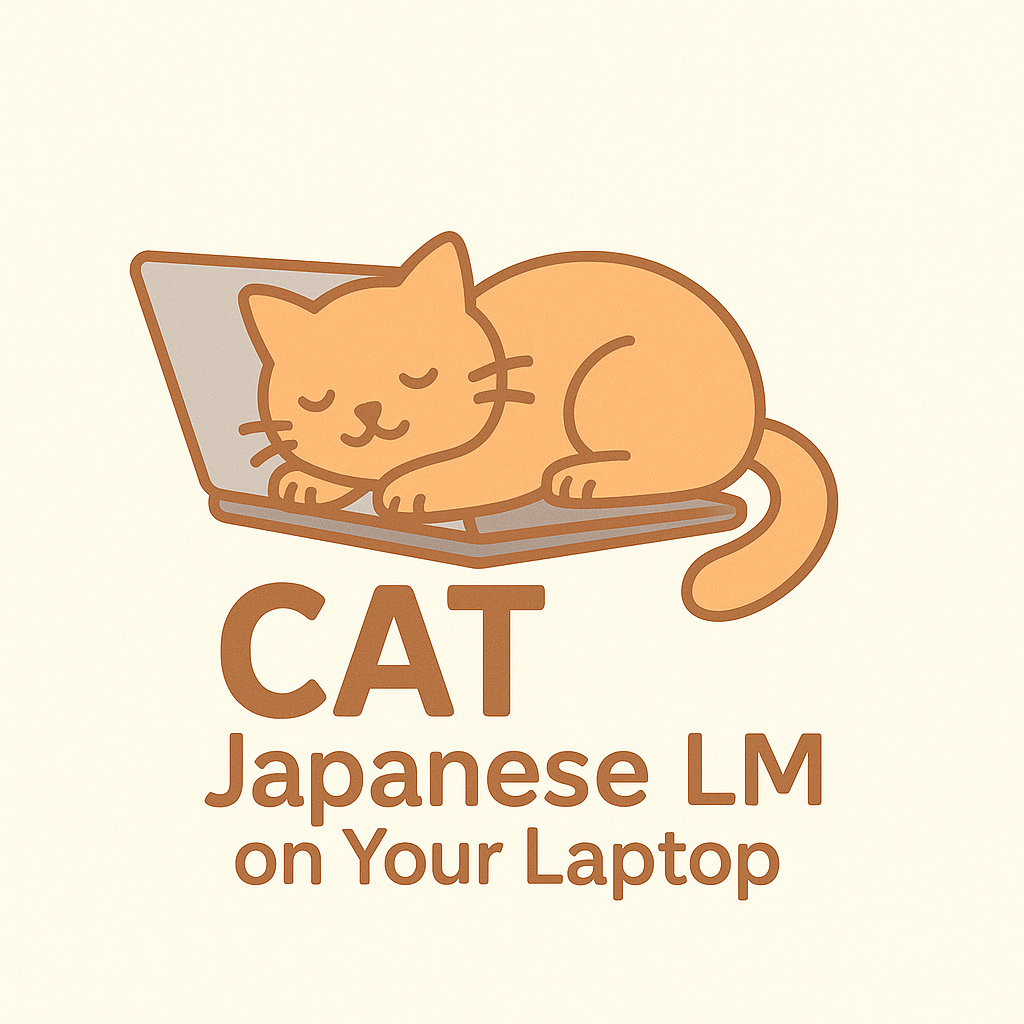}
    \captionsetup{labelformat=empty}
    \caption{CAT-Translate runs on a consumer GPU. It is open source that you can adopt for free.}
    \label{fig:cat}
\end{figure}

In many real-world scenarios in business, legal, medical, financial, and patent domains, you only have consumer GPUs and the data cannot be shared outside due to privacy concerns or regulatory requirements.
We focus on Japanese-English translation as a case study due to resource constraints. While many applications require multilingual translation capabilities, there are also many use cases where only a specific language pair is relevant to the specific practitioner. % (e.g., patent translation, legal document translation, and medical translation).

% These models are trained on A100 GPUs with a total training time of approximately 1 to 3 weeks, which is a reasonable amount of computational resources for a wider range of practitioners.
To investigate this question empirically, we develop a family of open source bilingual language models (0.8B, 1.4B, 3.3B, and 7B parameters) specialized for Japanese-English bidirectional translation. % using only open source resources and publicly available datasets.
We evaluate the models on real-world translation benchmarks across business, legal, medical, financial, and patent domains.
While the multilingual models have shown to achieve high accuracy in WMT general tasks~\citep{aharoni-etal-2019-massively,JMLR:v22:20-1307,kocmi-etal-2024-findings,cui-etal-2025-multilingual,kocmi-etal-2025-findings}, the bilingual models outperforms the multilingual models in the real-world translation benchmarks we evaluated.
The result suggests that language-specific models may be valuable for the coverage of the tasks while multilingual models can be sufficient for generic translation tasks, giving anecdotal evidence to support that language-specific translation models may be worth developing despite the rise of high-quality multilingual translation models.

\section{Training}
\label{sec:training}
Our training pipeline consists of three stages: (1) synthetic data generation and filtering, (2) two-stage supervised fine-tuning, and (3) Multi-Objective GRPO reinforcement learning \citep{ichihara2025mo}. Each stage involves specific design choices to maximize translation quality while managing computational resources effectively.

\begin{figure}
    \centering
    \includegraphics[width=0.8\columnwidth]{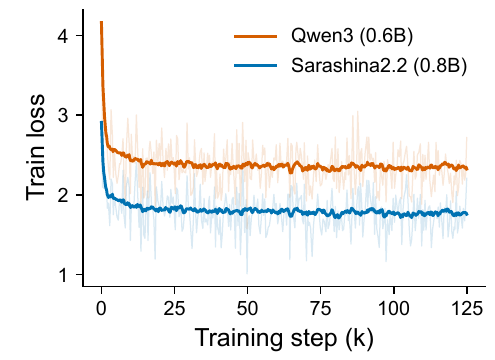} %{training_loss.png}
    \caption{Training loss curves for preliminary SFT experiments on Sarashina 2.2  (0.8B) and Qwen 3 (0.6B).}
    \label{fig:base-model}
\end{figure}

\subsection{Base Models}

We use the Sarashina 2.2 series\footnote{\url{https://huggingface.co/collections/sbintuitions/sarashina22}} of size 0.8B, 1.4B, and 3.3B parameters as our pretrained base models. These models are Japanese-English bilingual language models released under the MIT License, which aligns with our goal of providing fully open-source translation models.
In addition, we have a 7B in-house model as a base model. While the base model is not publicly available, the resulting translation model will be open source.

The selection of Sarashina 2.2 over higher-scoring alternatives like Qwen-3~\cite{yang2025qwen3} is driven by qualitative evaluation and with a preliminary experiment on SFT (Figure~\ref{fig:base-model}). While Qwen-3 demonstrated higher scores on standard benchmarks, manual inspection of generated translations revealed that Sarashina 2.2 produced more natural Japanese text that avoided {\it translationese}, unnatural phrasings that reflect source language structure rather than target language conventions \citep{Baker1993,freitag-etal-2020-bleu}. 
We tried to quantitatively evaluate the naturalness with automatic metrics such as COMET \citep{rei-etal-2022-comet} and LLM-as-a-judge \citep{kocmi-federmann-2023-large,kocmi-federmann-2023-gemba}, but the results were inconclusive. 
This suggests that naturalness is more difficult to learn through fine-tuning than translation accuracy, making it a more valuable property in the base model.

\subsection{Dataset}
We first describe the process of constructing our training dataset, which consists of synthetic data generated by large language models.

\paragraph{\bf Web-crawled parallel corpora are not sufficient.}
We have first evaluated the parallel corpora for Japanese and English, JParaCrawl \cite{morishita-etal-2020-jparacrawl,morishita-etal-2022-jparacrawl}, Laboro-ParaCorpus\footnote{\url{https://github.com/laboroai/Laboro-ParaCorpus}}, CCMatrix \cite{schwenk-etal-2021-ccmatrix}, NLLB \cite{nllb2024scaling}, and HPLT \cite{aulamo-etal-2023-hplt,burchell-etal-2025-expanded,obrien-etal-2025-dochplt}.
JParaCrawl and Laboro-ParaCorpus are parallel corpora specifically designed for Japanese-English translation, while CCMatrix, NLLB, and HPLT are multilingual corpora that include Japanese-English parallel data among many other language pairs.

We inspect some of the data entries and find that the quality of the parallel data is quite low for CCMatrix, NLLB, and HPLT, which are multilingual corpus that are not specifically designed for Japanese-English translation.
As a result, the majority of the data entries in these datasets contain significant translation errors that we consider to be low-quality for training modern translation models.
The quality of JParaCrawl and Laboro-ParaCorpus is significantly better than the others, but still not sufficient for training high-quality translation models.
Various filtering approaches to improve the quality of the parallel data are tested, but the resulting data are either low quality or too small, which is not ideal for training general-purpose translation models.
COMET-QE~\citep{rei-etal-2020-unbabels}\footnote{In the course of model training, we use codes, models, and datasets available for commercial purposes as the goal is to solve real-world problems that include industry and commercial usage. To this end we use Unbabel/wmt20-comet-qe-da model available by Apache 2.0. Non-commercial users may use more recent higher accruacy models such as Unbabel/wmt23-cometkiwi-da-xxl, which is published with non-commercial license.} showed some effectiveness in filtering out low-quality translations with high precision, but it was not effective in distinguishing translations for non-standard texts (e.g., slangs in Japanese), resulting in low recall biased filtering.
This would be useful if the target domain is restricted to formal texts, but it is not ideal for training general-purpose translation models that should be robust to various text styles and domains.

Overall, the resulting corpus from web-crawled datasets do not meet the quality and quantity requirements for training high-quality translation models, which motivated us to synthesize parallel data using large language models.

\paragraph{\bf Constructing monolingual corpora.}
Due to the limited availability of high-quality parallel corpora for specialized domains, we synthesize parallel data from monolingual sources using large language models. 
First we gathered monolingual corpora in Japanese and English from various sources including: in-house web-crawled data, fineweb~\citep{penedo2024the}, Laboro-ParaCorpus (the Japanese part), research abstracts from arXiv, PubMed, and J-Stage\footnote{\url{https://www.jstage.jst.go.jp}} (Japanese platform for academic journals), and patent documents from USPTO.\footnote{\url{https://www.uspto.gov}}
These sources provide a diverse range of text styles and domains, which is important for training robust translation models. The monolingual corpora are preprocessed to remove low-quality text and ensure a clean input for the data synthesis stage.

We first clean the monolingual corpora to remove undesirable texts from the dataset.
We investigate the quality of the monolingual data by manually inspecting random samples and find that the following filtering pipeline effectively removes undesirable text while retaining a diverse range of styles and domains.
The filtering pipeline of the monolingual data consists of the following steps:
\begin{enumerate}
    \item \textbf{Language filtering}: Remove texts written predominantly in languages other than Japanese and English using \texttt{fastText} language identification~\citep{joulin-etal-2017-bag}. We find that \texttt{fastText} retains some non-Japanese text (e.g., Simplified Chinese, Traditional Chinese, and Korean) in the Japanese monolingual corpus. We look for tools to identify these automatically including \texttt{langid} \cite{lui-baldwin-2012-langid}, \texttt{lingua-py}\footnote{\url{https://github.com/pemistahl/lingua-py}}, \texttt{Compact Language Detector v3}\footnote{\url{https://github.com/google/cld3}}, and \texttt{language-detection}\footnote{\url{https://github.com/shuyo/language-detection}}. However, we do not find significant improvement on distinguishing these instances using these libraries. We assume that inclusion of some non-Japanese text in the {\it source text} is unlikely to cause significant issues in the training, so these data entries are retained.  
    
    \item \textbf{Absolute length filtering}: Remove instances that are too long to fit in the context window of the base models (i.e., 4096 tokens) and instances shorter than three words. We aim to train models that can handle both long and short inputs and outputs, including sentence-length, paragraph-length, and multiple-paragraph-length translations. On the other hand, the base models we use are small and their context length is limited to 8192 tokens. Therefore, we filter out instances that are too long to fit in the context window. We also filter out instances shorter than three words as we find that many corpus entries with only one or two words often have some bias and duplicates (e.g., generated by some template). Because the following MinHash-based deduplication step is not that effective in removing these short instances, we apply this length filtering as a heuristic. In our experiments, we do not find the resulting model to be worse in handling short inputs.
    
    \item \textbf{Deduplication}: Apply MinHash-based near-duplicate detection using Duplodocus\footnote{\url{https://github.com/allenai/duplodocus}} to remove redundant training instances \citep{lee-etal-2022-deduplicating}.
    We find that the deduplication step significantly reduced the number of near-duplicate instances in the training data.
    Although it is ideal to evaluate the effect of the deduplication on the final translation quality, it requires significant computational resources to train models with and without the deduplication step, which we cannot afford under our resource constraints. However, we find that the deduplication step significantly reduced the number of near-duplicate instances in the training data, which we expect to improve the quality of the trained models. The hyperparameters for the deduplication step are tuned manually. We keep track of several sets of instances that are near-duplicates and not duplicates. Hyperparameters that remove the near-duplicate sets while retaining the non-duplicate sets are selected.
\end{enumerate}

\paragraph{\bf Synthesizing parallel data.}
We first used DeepSeek-R1~\citep{guo2025deepseek} for initial prototyping of our data synthesis pipeline, and find that the quality of the generated translations was sufficient for many instances. However, DeepSeek-R1 requires a significant amount of computational resources that we cannot afford under our resource constraints. 
Therefore, we switch to gpt-oss-20b~\citep{openai2025gptoss} for the majority of our data synthesis, which provids a good balance of quality and efficiency. 
However, we find that gpt-oss-20b struggles to translate some challenging inputs such as research abstracts in PubMed with multiple technical terms. These domains are also challenging to evaluate manually as authors of the paper are not experts in these domains, so we use gpt-oss-120b~\citep{openai2025gptoss} to generate translations for these challenging inputs to ensure higher translation quality in these areas.

We apply a multi-stage filtering pipeline to the generated parallel data:
\begin{enumerate}
    \item \textbf{Length ratio filtering}: Filter instances where the ratio of Japanese to English text lengths falls outside acceptable bounds, as extreme ratios often indicate translation errors~\citep{hoang-koehn-2008-design}. We set the acceptable range for Japanese characters / English characters ratio to [0.5, 2.0]. This filtering is conservative to only remove outliers while retaining a wide variety of sentence structures and styles, including those with significant length differences between Japanese and English.
    We find that this length ratio filtering is effective in removing low-quality translations, especially those with significant omissions or additions (e.g., adding explanation to the translation).
    
    \item \textbf{Rule-based filtering}: Apply hand-crafted rules to detect and remove common error patterns identified during manual review. 
    Even though gpt-oss models generate high-quality translations, we find that they still exhibit some common error patterns that can be effectively filtered with simple rules. 

    For example, we found that gpt-oss models often generate markdown-formatted outputs (e.g., using \texttt{*} for bullet points and \texttt{\#} for headings) even when the input is plain text. 
    We first try using the lexer guesser of {\it Pygments}\footnote{\url{https://github.com/pygments/pygments}} library to automatically detect the format language of the generated translation, but because the texts are often a mixture of plain text and markdown, the lexer guesser is not effective in distinguishing these instances. It is also the case that some markdown commands seem appropriate for English text but not for the corresponding Japanese text, which makes it difficult to apply the lexer guesser to the generated translation as a whole. 
    Therefore, we instead remove instances where the generated translation are significantly different in terms of formatting compared to the source text.    
    We apply a simple rule-based filter that counts the number of markdown formatting characters in the generated translation and removes instances that have more than twice the number of such characters in the source text.

    Another error pattern is that gpt-oss models sometimes fail to finish the reasoning trace, resulting in incomplete translations. We apply a simple rule-based filter that removes instances where the generated response contains the special characters gpt-oss use for reasoning. Due to a human error, some instances with incomplete reasoning traces are included in the training data. We find the model to occasionally start reasoning instead of translating the input, which is likely due to the presence of these incomplete reasoning traces in the training data.

    gpt-oss occasionally refuses to translate instances that they identify as censored content, often from PubMed abstracts~\citep{rottger-etal-2024-xstest,cui2025orbench}. Because the refusal response are often much shorter than expected, the length ratio filtering step effectively removes most of these instances. To make sure, we apply a simple keyword-based filter that identifies refusal responses if certain keywords (e.g., ``refuse'', ``sorry'', ``censored'', etc.) are present multiple times. Many of the identified instances are not refusal responses, but for higher precision, we decided to remove all the identified instances. This might introduce some bias but PubMed abstracts are large enough that removing some of the instances will unlikely to affect the quality of the data significantly.
\end{enumerate}

Once the filtering is applied, we inspect random samples to verify the quality of the resulting parallel data.
We find the quality of the parallel data to be good enough for real-world usage.
To increase the quantity of the parallel data, we generate additional parallel data by back-translation using the same gpt-oss models. 
We use the same filtering pipeline for the back-translated data, and we find that the quality of the back-translated data is comparable to the forward-translated data except for scientific abstracts and patent documents where some generations have noticeable translation errors. We decide not to use any of back-translated data for scientific abstracts.
Note that we only use instances generated by translating English source text to Japanese target text to train for English-to-Japanese translation. We do not use instances generated by translating Japanese source text to English target text to train for English-to-Japanese translation, and vice versa. This is common in Japanese-English translation~\citep{hirano2025plamo}.
% At least for Japanese-English language pair, translation is not symmetric and the translation quality can be significantly different between the two directions.

\begin{figure}
    \centering
    \includegraphics[width=0.8\columnwidth]{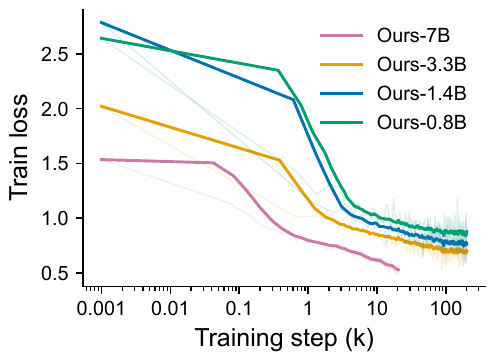} %{dataset_composition.png}
    \caption{Training loss curves for the Stage 1 of supervised fine-tuning.}
    \label{fig:sft1-loss}
\end{figure}

\begin{figure}
    \centering
    \includegraphics[width=0.8\columnwidth]{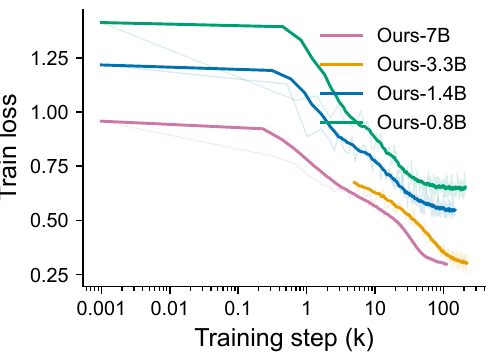} %{dataset_composition.png}
    \caption{Training loss curves for the Stage 2 of supervised fine-tuning. The training of 3.3B model has crashed once and restarted, which is the reason for the discontinuity in the loss curve.}
    \label{fig:sft2-loss}
\end{figure}
\subsection{Two-Stage Supervised Fine-Tuning (SFT)}

We employ a two-stage fine-tuning approach that balances diversity and quality in the training data.
The training loss curves of the two stages are shown in Figures~\ref{fig:sft1-loss} and \ref{fig:sft2-loss}.

\paragraph{\bf Stage 1: Diversity focus.}
The first stage prioritizes exposure to diverse translation scenarios. The training corpus consists primarily of web-crawled data that are relatively tolerant in the output variety, supplemented by domains which tolerate little variations such as scientific abstracts (arXiv and PubMed) and patents (USPTO). 
Most instances are sentence-length, with some paragraph-length examples included.
During the course of training, the model performance largely saturated at approximately 100k training steps (Figure~\ref{fig:sft1-loss}). This saturation motivates our second-stage approach, which focuses on quality over quantity.

\paragraph{\bf Stage 2: Quality focus.}
We investigate the quality of the trained models after the first stage and find that the models perform reasonably well on general translation tasks, but struggle with more challenging inputs such as scientific abstracts and patents.
The second stage emphasizes high-quality translations for challenging inputs. A large portion of the data is generated by gpt-oss-120b rather than the smaller gpt-oss-20b, ensuring higher translation quality. The training corpus focuses on research abstracts, patent documents, and underspecified or misspecified text including inputs with typos and ambiguous phrasing.
Most instances in this stage are paragraph-length to multiple-paragraph-length, requiring the model to maintain coherence and context across longer inputs. 
We retain 10\% of the instances from the first stage to maintain diversity of the inputs.

\begin{figure*}
    \centering
    \includegraphics[width=0.95\textwidth]{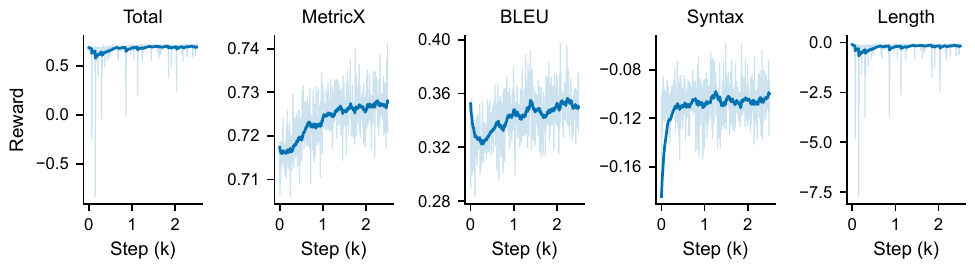} %{dataset_composition.png}
    \caption{The curves for the Multi-Objective GRPO reward values. The training of 1.4B model is shown as all the other runs have at least one discontinuity due to hardware issues (e.g., CUDA out of memory). The other models show similar trends.}
    \label{fig:mogrpo-loss}
\end{figure*}

\subsection{Multi-Objective Group Relative Policy Optimization (MO-GRPO)}

The models trained through two stages of SFT achieve reasonable translation quality. 
They generate sufficiently good translations and the fluency and naturalness are high for both English and Japanese.
However, they often make non-critical but noticeable mistakes that can be easily identified by human evaluators, which makes the system less useful in real-world applications where trust from human users is important.

We apply Multi-Objective Group Relative Policy Optimization (MO-GRPO; \cite{ichihara2025mo}) to further improve translation quality through reinforcement learning. 
MO-GRPO is a reinforcement learning algorithm designed for language generation tasks that optimizes multiple reward objectives simultaneously while maintaining stable training dynamics.
We use it to optimize a set of lightweighted reward functions at the same time to mitigate the weaknesses of any single reward function without the use of computationally expensive reward models such as DeepSeek-R1.
The training dataset is the rest of the data from Stage 2 of SFT.

\paragraph{\bf Reward model design.}
The reward model is a critical component of the reinforcement learning stage, as it provides the feedback signal that guides the model's learning process. We design a composite reward function that combines multiple components to address different aspects of translation quality and mitigate potential weaknesses in any single metric.
We use the following reward components:
\begin{itemize}
    \item \textbf{MetricX-24}, a learned reference-free quality estimation metric with high correlation to human judgments~\citep{juraska-etal-2024-metricx}.
    \item \textbf{BLEU}, a traditional n-gram overlap metric that provides an independent quality signal and rewards accurate translation of specific terms~\citep{papineni-etal-2002-bleu}.
    \item \textbf{Format consistency penalty}, an absolute penalty for translations that exhibit significant format differences from the source text, addressing the tendency for models to generate markdown-formatted outputs.
    \item \textbf{Length penalty}, an absolute penalty for translations that are excessively long or short relative to the source text and reference translation, effectively suppressing hallucinations.
\end{itemize}
MetricX-24 is our primary reward model~\citep{juraska-etal-2024-metricx}. MetricX-24 is one of the most accurate metric for machine translation available that predicts human judgments (MQM) on the quality of the translations. We choose it as it is open source (Apache 2.0), efficient to compute, and has demonstrated high correlation with human judgments in previous evaluations \citep{juraska-etal-2024-metricx,freitag-etal-2024-llms}. Translategemma~\citep{finkelstein2026translategemma} uses an ensemble of reward models including MetricX for reinforcement learning, so it is likely a promising approach to use it.
Because MetricX-24 outputs estimated MQM score from 0 to 25 where 0 is the best score and 25 is the worst score, we convert it to a reward value by negating the score and applying a linear transformation to scale it to a range of [0, 1], where $1$ corresponds to the best possible translation and $0$ corresponds to the worst possible translation.
We use an XL model of MetricX-24\footnote{\url{https://huggingface.co/google/metricx-24-hybrid-xl-v2p6-bfloat16}} as in our computational environment XXL model does not fit in the VRAM efficiently under our configuration, and the two models show similar accuracy in our preliminary experiments.

\begin{table*}[t]
\centering
% \resizebox{\textwidth}{!}{%
\begin{tabular}{lrrr}
\toprule
 & & \multicolumn{1}{c}{Ja$\rightarrow$En} & \multicolumn{1}{c}{En$\rightarrow$Ja} \\
\cmidrule(lr){3-3}\cmidrule(lr){4-4}
Model & Avg. & WMT21 & WMT24Doc \\
\midrule
pfnet/plamo-2-translate                           & 4.393 & 4.413 & 4.373 \\
tencent/HY-MT1.5-7B                               & 3.673 & 3.394 & 3.953 \\
shisa-ai/shisa-v2.1-llama3.2-3b                   & 3.486 & 2.818 & 4.154 \\
Unbabel/Tower-Plus-9B                             & 3.311 & 3.568 & 3.053 \\
google/translategemma-12b-it                      & 3.285 & 2.842 & 3.728 \\
nvidia/NVIDIA-Nemotron-Nano-9B-v2-Japanese        & 3.248 & 3.407 & 3.089 \\
{\bf CAT-Translate-1.4B}                                   & 3.191 & 2.985 & 3.396 \\
microsoft/phi-4                                   & 3.187 & 3.515 & 2.858 \\
{\bf CAT-Translate-7B}                                     & 3.143 & 3.356 & 2.929 \\
Qwen/Qwen2.5-14B-Instruct                         & 3.106 & 2.868 & 3.343 \\
shisa-ai/shisa-v2.1-lfm2-1.2b                     & 3.103 & 2.366 & 3.840 \\
{\bf CAT-Translate-3.3B}                                   & 3.057 & 3.286 & 2.828 \\
tokyotech-llm/Llama-3.1-Swallow-8B-Instruct-v0.5  & 3.049 & 1.849 & 4.249 \\
google/translategemma-4b-it                       & 2.872 & 2.636 & 3.107 \\
tencent/HY-MT1.5-1.8B                             & 2.847 & 2.783 & 2.911 \\
{\bf CAT-Translate-0.8B}                                   & 2.745 & 2.672 & 2.817 \\
sbintuitions/sarashina2.2-3b-instruct-v0.1        & 2.715 & 2.696 & 2.734 \\
mistralai/Ministral-8B-Instruct-2410              & 2.676 & 2.501 & 2.852 \\
SakanaAI/TinySwallow-1.5B-Instruct                & 2.490 & 2.424 & 2.556 \\
LiquidAI/LFM2.5-1.2B-JP                          & 2.399 & 2.260 & 2.538 \\
google/gemma-2-2b-jpn-it                          & 2.370 & 2.398 & 2.343 \\
LiquidAI/LFM2-350M-ENJP-MT                        & 2.362 & 2.689 & 2.036 \\
llm-jp/llm-jp-3.1-1.8b-instruct4                  & 2.291 & 2.600 & 1.982 \\
Rakuten/RakutenAI-2.0-mini-instruct               & 1.804 & 1.785 & 1.822 \\
meta-llama/Llama-3.2-3B-Instruct                  & 1.667 & 1.850 & 1.485 \\
\bottomrule
\end{tabular}%
% }
\caption{M-Prometheus scores on WMT benchmarks. Scores are on a 1--5 scale. WMT21 is the WMT21 test set (Ja$\rightarrow$En) and WMT24Doc is the WMT24 document-level test set (En$\rightarrow$Ja). While we try not to use it extensively, CAT-Translate models use these tests as the validation set on the course of development, which may introduce overoptimization to these benchmark scores.}
\label{tab:wmt}
\end{table*}

However, MetricX-24 has several limitations that can be exploited by generation models.
First, MetricX-24 is inherently multilingual, assigning scores regardless of output language, potentially rewarding outputs that fail to translate into the target language (e.g., you get the best score by responding the source text as is in the source language).
Second, it is format-agnostic, largely ignoring syntactic formatting characters such as newlines and markdown syntax (e.g., \texttt{*} and \texttt{\#}), which can lead to unnatural translations that are still rewarded by the metric.
Third, we find MetricX-24 to be relatively tolerant of outputs with additional explanation of the translation, which is undesirable for a stand alone translation model that should only output the translation without additional commentary.
We address these concerns with the following auxiliary rewards.

\paragraph{\bf BLEU score.}
We compute BLEU scores~\citep{papineni-etal-2002-bleu} against reference translations to measure lexical overlap using SacreBLEU~\citep{post-2018-call}. This serves two purposes: (1) avoiding over-optimization to MetricX-24 by providing an independent quality signal, and (2) rewarding accurate translation of technical terms and specific word choices that may be underweighted by learned metrics.
Over-optimization to learned metrics is a well-known issue in reinforcement learning for language generation \citep{Goodhart1984,pan2022the,pmlr-v202-gao23h}, and the inclusion of a traditional n-gram overlap metric like BLEU helps mitigate this risk by providing a complementary signal that is less susceptible to exploitation \citep{pombal-etal-2025-adding}.

\paragraph{\bf Format consistency penalty ($\text{FmtDist}$).}
We penalize translations that exhibit significant format differences from the source text. 
This addresses the observed tendency for models to generate markdown-formatted outputs even when the input is plain text. 
We extract the sequence of formatting characters (e.g., newlines, \texttt{*}, \texttt{\#}, etc.) from both the source text and the generated translation. Then, we compute $\text{FmtDist}(x, y)$, the edit distance between these two sequences where the operation cost for insertion, deletion, substitution, and transposition is 1. We penalize generations with an edit distance larger than 5, which we find to be effective in keeping the generated translations to be in a similar format as the source text. If the edit distance is within 15, a linear penalty is applied based on the edit distance. If the edit distance is larger than 15, the entire reward value $R(x, y, y^*)$ is set to zero.

\paragraph{\bf Length penalty ($\text{Length}$).}
We penalize translations that are excessively long or short relative to the source text and reference translation. This constraint effectively suppressed hallucinations where models added extraneous information.
If the length of the generated translation is within (0.8, 1.2) times the length of the reference translation, no penalty is applied. If within (0.5, 2.0), a linear penalty is applied based on the deviation from the acceptable range. If outside (0.5, 2.0), the entire reward value $R(x, y, y^*)$ is set to zero.

The resulting reward function is as follows:
\begin{align*}
    R(x, y, y^*) &= \mathrm{norm}(\text{MetricX-24}(x, y, y^*))\\
            &+ 0.1 \cdot \mathrm{norm}(\text{BLEU}(y, y^*))\\
            &- \lambda_1\, \text{FmtDist}(x, y) \\
            &- \lambda_2\, \text{Length}(y, y^*),
\end{align*}
where $x$ is the source text, $y$ is the generated translation, and $y^*$ is the reference translation. The $\lambda_1$ and $\lambda_2$ are hyperparameters that control the strength of the format consistency penalty and length penalty, respectively. We set $R$ to be non-negative and set $\lambda_1$ and $\lambda_2$ to large values to ensure the reward is zero when the constraints are violated.

% We apply different normalization strategies to different reward components.
We normalize MetricX-24 and BLEU to compute relative advantages within each batch for each reward component, following the approach of MO-GRPO~\citep{ichihara2025mo}. This allows the model to learn from relative improvements in these metrics, which is important for guiding learning in a way that is robust to scale differences and potential exploitation. Because MetricX-24 has higher correlation with human judgments, we assign it a higher weight in the reward function, while BLEU serves as a complementary signal to prevent over-optimization to MetricX-24.

Format and length penalties are applied as absolute values without normalization, similar to Dr.~GRPO~\citep{liu2025understanding}. The rationale is that these constraints are straightforward to learn and should be enforced regardless of translation quality improvements. Large absolute penalties prevent the model from violating these constraints even when doing so might improve translation quality.

Figure~\ref{fig:mogrpo-loss} shows the reward values during the course of MO-GRPO training. We observe the reward values to be increasing during the course of training, which indicates that the model is learning to improve the translation quality according to the reward function.
We also monitor the generated texts manually to see if the model is improving in terms of translation quality and if there are any signs of degeneration or exploitation of the reward function.
As far as we are aware of, the reward model we designed is not being exploited by the model, and the translation quality is improving during the course of training without any signs of degeneration.

\begin{table*}[t]
\centering
\resizebox{\textwidth}{!}{%
\begin{tabular}{lrrrrrrrrrrrrr}
\toprule
 & & \multicolumn{6}{c}{Ja$\rightarrow$En} & \multicolumn{5}{c}{En$\rightarrow$Ja} \\
\cmidrule(lr){3-8}\cmidrule(lr){9-13}
Model & Avg. & Avg. & BSD & Court & JMed & PFMT & PAT & Avg. & BSD & JMed & PFMT & PAT \\
\midrule
{\bf CAT-Translate-7B}                       & 37.68 & 41.06 & 33.75 & 45.29 & 30.65 & 49.86 & 45.74 & 34.31 & 16.29 & 29.62 & 52.94 & 38.37 \\
{\bf CAT-Translate-3.3B}                     & 36.16 & 37.51 & 26.51 & 42.44 & 24.47 & 49.93 & 44.23 & 34.80 & 17.21 & 28.67 & 53.88 & 39.44 \\
{\bf CAT-Translate-1.4B}                     & 33.73 & 33.26 & 31.28 & 43.84 & 24.08 & 36.55 & 30.57 & 34.19 & 15.71 & 26.92 & 51.53 & 42.58 \\
Unbabel/Tower-Plus-9B                             & 32.41 & 36.84 & 15.43 & 40.54 & 29.13 & 58.00 & 41.10 & 27.99 & 10.00 & 18.80 & 53.00 & 30.16 \\
google/translategemma-12b-it                      & 32.24 & 35.81 & 31.58 & 34.30 & 23.46 & 48.75 & 40.97 & 28.68 & 15.92 & 21.79 & 52.53 & 24.47 \\
{\bf CAT-Translate-0.8B}                     & 30.42 & 29.71 & 29.63 & 33.19 & 22.96 & 32.51 & 30.56 & 30.68 & 14.60 & 26.22 & 50.62 & 32.87 \\
google/translategemma-4b-it                       & 28.09 & 29.41 & 28.86 & 25.89 & 21.50 & 42.65 & 28.16 & 26.76 & 14.14 & 20.68 & 51.99 & 20.23 \\
LiquidAI/LFM2.5-1.2B-JP                          & 25.47 & 24.51 & 19.06 & 29.99 & 22.10 & 43.61 &  7.80 & 26.43 & 14.57 & 23.85 & 54.77 & 12.54 \\
pfnet/plamo-2-translate                           & 25.24 & 25.92 & 25.55 & 28.63 & 22.90 & 29.02 & 23.48 & 24.57 & 17.35 & 24.98 & 32.04 & 23.89 \\
LiquidAI/LFM2-350M-ENJP-MT                        & 24.95 & 24.91 & 10.94 & 29.56 & 21.48 & 41.40 & 21.17 & 25.00 &  8.11 & 22.84 & 47.53 & 21.52 \\
mistralai/Ministral-8B-Instruct-2410              & 24.12 & 27.52 & 19.23 & 29.21 & 16.25 & 50.23 & 22.69 & 20.71 & 12.91 & 16.49 & 41.66 & 11.80 \\
nvidia/NVIDIA-Nemotron-Nano-9B-v2-Japanese        & 22.97 & 22.77 &  9.62 & 34.98 & 18.01 & 38.44 & 12.81 & 23.18 & 10.62 & 20.41 & 42.55 & 19.13 \\
Rakuten/RakutenAI-2.0-mini-instruct               & 18.43 & 17.24 &  0.11 & 30.62 & 18.21 & 29.34 &  7.90 & 19.62 &  5.19 & 20.36 & 45.70 &  7.23 \\
SakanaAI/TinySwallow-1.5B-Instruct                & 15.74 & 14.99 &  4.96 & 18.93 & 15.83 & 26.67 &  8.58 & 16.49 &  6.30 & 17.58 & 34.07 &  8.00 \\
llm-jp/llm-jp-3.1-1.8b-instruct4                  & 15.18 & 16.26 & 18.82 &  2.44 & 15.67 & 30.65 & 13.72 & 14.11 & 15.38 &  4.91 & 25.47 & 10.65 \\
tencent/HY-MT1.5-1.8B                             & 14.49 &  8.95 &  5.50 &  4.59 &  4.00 & 15.67 & 14.98 & 20.04 &  6.33 & 18.13 & 37.75 & 17.96 \\
shisa-ai/shisa-v2.1-llama3.2-3b                   & 14.27 & 14.26 & 17.08 &  3.70 &  8.26 & 26.86 & 15.42 & 14.28 & 13.18 &  5.54 & 25.97 & 12.41 \\
google/gemma-2-2b-jpn-it                          & 14.15 & 16.98 & 20.04 &  8.08 & 11.27 & 31.49 & 14.01 & 11.32 & 12.37 &  4.48 & 16.24 & 12.21 \\
shisa-ai/shisa-v2.1-lfm2-1.2b                     & 13.08 & 14.02 & 20.93 &  4.95 &  7.68 & 26.72 &  9.80 & 12.14 & 12.11 &  5.54 & 17.60 & 13.30 \\
microsoft/phi-4                                   & 11.92 & 13.48 &  6.10 & 18.66 &  2.81 & 24.86 & 14.98 & 10.36 &  3.24 &  6.97 & 14.36 & 16.87 \\
tencent/HY-MT1.5-7B                               & 10.56 & 13.46 &  4.99 & 12.32 &  5.72 & 29.53 & 14.76 &  7.67 &  0.82 &  7.80 & 14.30 &  7.74 \\
tokyotech-llm/Llama-3.1-Swallow-8B-Instruct-v0.5  & 10.35 & 12.42 & 24.25 &  2.30 &  3.69 & 14.11 & 17.74 &  8.28 &  6.82 &  2.37 & 11.21 & 12.71 \\
Qwen/Qwen2.5-14B-Instruct                         &  8.39 &  9.88 & 10.81 &  4.70 &  4.27 & 11.18 & 18.46 &  6.89 &  4.01 &  3.69 & 13.42 &  6.42 \\
meta-llama/Llama-3.2-3B-Instruct                  &  6.06 &  9.90 & 18.60 &  0.41 &  2.72 & 16.62 & 11.17 &  2.23 &  1.44 &  1.10 &  4.50 &  1.87 \\
\bottomrule
\end{tabular}%
}
\caption{BLEU scores on real-world translation benchmarks. The Court benchmark does not have En$\rightarrow$Ja direction. The models are sorted by the macro average score.}
\label{tab:bleu}
\end{table*}
\begin{table*}[t]
\centering
\resizebox{\textwidth}{!}{%
\begin{tabular}{lrrrrrrrrrrrrr}
\toprule
 & & \multicolumn{6}{c}{Ja$\rightarrow$En} & \multicolumn{5}{c}{En$\rightarrow$Ja} \\
\cmidrule(lr){3-8}\cmidrule(lr){9-13}
Model & Avg. & Avg. & BSD & Court & JMed & PFMT & PAT & Avg. & BSD & JMed & PFMT & PAT \\
\midrule
shisa-ai/shisa-v2.1-llama3.2-3b                   & 4.589 & 4.733 & 4.559 & 4.440 & 4.842 & 4.900 & 4.923 & 4.409 & 4.114 & 4.494 & 4.571 & 4.455 \\
pfnet/plamo-2-translate                           & 4.525 & 4.602 & 4.676 & 4.440 & 4.657 & 4.700 & 4.538 & 4.429 & 4.486 & 4.514 & 4.714 & 4.000 \\
tokyotech-llm/Llama-3.1-Swallow-8B-Instruct-v0.5  & 4.501 & 4.644 & 4.676 & 4.640 & 4.581 & 4.400 & 4.923 & 4.323 & 4.286 & 4.265 & 4.286 & 4.455 \\
shisa-ai/shisa-v2.1-lfm2-1.2b                     & 4.370 & 4.427 & 4.265 & 4.720 & 4.845 & 5.000 & 3.308 & 4.298 & 4.143 & 4.385 & 4.571 & 4.091 \\
google/translategemma-12b-it                      & 4.264 & 4.536 & 4.441 & 4.780 & 4.803 & 4.500 & 4.154 & 3.924 & 4.257 & 4.153 & 4.286 & 3.000 \\
{\bf CAT-Translate-7B}                                     & 4.084 & 4.418 & 3.971 & 4.880 & 4.777 & 5.000 & 3.462 & 3.666 & 3.743 & 4.063 & 3.857 & 3.000 \\
{\bf CAT-Translate-1.4B}                                   & 4.034 & 4.148 & 3.294 & 4.700 & 4.748 & 5.000 & 3.000 & 3.891 & 4.029 & 4.139 & 4.214 & 3.182 \\
google/translategemma-4b-it                       & 3.929 & 4.297 & 4.000 & 4.600 & 4.662 & 4.300 & 3.923 & 3.469 & 4.057 & 3.742 & 3.714 & 2.364 \\
{\bf CAT-Translate-3.3B}                                   & 3.730 & 4.082 & 3.735 & 4.840 & 4.803 & 4.800 & 2.231 & 3.290 & 3.200 & 4.066 & 3.714 & 2.182 \\
{\bf CAT-Translate-0.8B}                                   & 3.657 & 3.876 & 2.735 & 4.720 & 4.661 & 4.800 & 2.462 & 3.384 & 3.029 & 3.956 & 3.643 & 2.909 \\
Qwen/Qwen2.5-14B-Instruct                         & 3.545 & 3.736 & 4.412 & 3.940 & 3.380 & 3.100 & 3.846 & 3.308 & 3.514 & 3.690 & 4.571 & 1.455 \\
microsoft/phi-4                                   & 3.342 & 3.558 & 4.206 & 3.200 & 2.792 & 4.900 & 2.692 & 3.072 & 3.171 & 2.913 & 3.929 & 2.273 \\
google/gemma-2-2b-jpn-it                          & 3.241 & 3.657 & 3.882 & 3.160 & 4.075 & 4.400 & 2.769 & 2.721 & 2.943 & 2.342 & 3.143 & 2.455 \\
LiquidAI/LFM2.5-1.2B-JP                          & 3.172 & 3.310 & 2.000 & 4.440 & 4.355 & 4.600 & 1.154 & 3.000 & 3.057 & 3.702 & 3.786 & 1.455 \\
mistralai/Ministral-8B-Instruct-2410              & 3.147 & 3.695 & 3.794 & 4.000 & 3.698 & 4.600 & 2.385 & 2.462 & 2.486 & 2.868 & 2.857 & 1.636 \\
SakanaAI/TinySwallow-1.5B-Instruct                & 3.036 & 3.239 & 2.647 & 3.780 & 3.820 & 4.100 & 1.846 & 2.782 & 2.886 & 3.138 & 3.286 & 1.818 \\
LiquidAI/LFM2-350M-ENJP-MT                        & 2.816 & 3.100 & 1.353 & 4.060 & 4.003 & 4.700 & 1.385 & 2.461 & 1.314 & 3.290 & 3.786 & 1.455 \\
llm-jp/llm-jp-3.1-1.8b-instruct4                  & 2.355 & 2.633 & 1.941 & 2.080 & 3.660 & 4.100 & 1.385 & 2.008 & 2.086 & 1.686 & 2.714 & 1.545 \\
Rakuten/RakutenAI-2.0-mini-instruct               & 2.222 & 2.344 & 1.147 & 2.940 & 3.001 & 3.400 & 1.231 & 2.069 & 1.257 & 2.805 & 3.214 & 1.000 \\
tencent/HY-MT1.5-1.8B                             & 2.109 & 1.726 & 2.353 & 1.300 & 1.567 & 2.100 & 1.308 & 2.588 & 2.000 & 3.136 & 3.214 & 2.000 \\
tencent/HY-MT1.5-7B                               & 2.074 & 2.355 & 1.588 & 2.460 & 1.825 & 3.900 & 2.000 & 1.723 & 1.143 & 1.846 & 2.357 & 1.545 \\
meta-llama/Llama-3.2-3B-Instruct                  & 1.837 & 2.373 & 3.000 & 1.300 & 2.402 & 3.700 & 1.462 & 1.166 & 1.171 & 1.351 & 1.143 & 1.000 \\
\bottomrule
\end{tabular}%
}
\caption{M-Prometheus scores on real-world translation benchmarks. Scores are on a 1--5 scale. The Court benchmark does not have En$\rightarrow$Ja direction. The models are sorted by the macro average score.}
\label{tab:prometheus}
\end{table*}

\begin{table*}[t]
\centering
\adjustbox{max width=0.95\textwidth}{
% \begin{tabular}{@{}lll@{}}
{\footnotesize
\begin{tabular}{p{0.15\textwidth}p{0.7\textwidth}p{0.1\textwidth}}
\toprule
Source (Japanese) & \JPN{ヒドロコルチゾン＋フルドロコルチゾンによる治療、またはドロトレコジンα（活性型）（drotrecogin alfa [activated]）による治療は、宿主反応を調節できることから、敗血症性ショック患者の臨床転帰を改善するという仮説を立てた。} &  Score$\uparrow$ (Prometheus) \\
\midrule
Reference (English) & We hypothesized that therapy with hydrocortisone plus fludrocortisone or with drotrecogin alfa (activated), which can modulate the host response, would improve the clinical outcomes of patients with septic shock. \\
\midrule
Model Output (HY-MT1.5-7B) &  Assistant...  
PROFESSOR: Professor...  
We decided to test this hypothesis through clinical trials.  
{\color{red} We conducted clinical trials to evaluate the efficacy of these treatments in reducing mortality and mortality rates.  
We conducted clinical trials to evaluate the effectiveness of these treatments in reducing mortality and mortality rates. (...Repetition until max token length)} & 1 \\
\midrule
Model Output (shisa-v2.1-llama3.2-3b) & \JPN{Treatment with hydrocortisone + dexamethasone, or drotrecogin alfa (active) (drotrecogin alfa [activated]) has been hypothesized to improve clinical outcomes in septic shock patients due to its ability to modulate the host response.  

{\color{red}(Note: ""宿主反応"" is translated as ""host response"" to maintain medical accuracy, as it refers to the body's immune and inflammatory reactions in the context of sepsis. ""ドロトレコジンα（活性型）"" is translated as ""drotrecogin alfa (active)"" to clarify the activated form of the drug, as ""drotrecogin alfa"" is the inactive precursor, and ""drotrecogin alfa [activated]"" is the active form used in clinical settings.)}} & {\color{red} 5} \\
\midrule
Model Output (CAT-Translate-7B) & We hypothesized that treatment with hydrocortisone plus fludrocortisone or drotrecogin alfa (activated) could improve clinical outcomes in patients with septic shock by modulating host responses. & 5 \\
\bottomrule
\end{tabular}
}
}
\caption{Translation example from the JMedBench.}
\label{tab:hy}
\end{table*}

\subsection{Validation on WMT}
\label{sec:validation}

During the course of training, we periodically evaluate the performance of the models on WMT test sets for sanity check. We use WMT 2021 Japanese-English and WMT 2024 Document-Level English-Japanese test sets for validation. These datasets are less popular for evaluating Japanese LLMs than the other WMT test sets, so they are less likely to be overfitted.
For the evaluation metric we use M-Prometheus-14B, which is an LLM-as-a-judge based metric with source, reference, and translation inputs developed by Unbabel \citep{pombal2025mprometheussuiteopenmultilingual,freitag-etal-2024-llms}. M-Prometheus is trained and evaluated in a variety of domains and text styles (Appendix A.3.2. in \cite{pombal2025mprometheussuiteopenmultilingual}), thus we expect it to be a good metric for validating the performance of our models.
The results of the final models are shown in Table~\ref{tab:wmt}.

Due to the limited computational resources, we do not tune hyperparameters based on the validation performance. We use it mainly for sanity check to make sure that the training is progressing in the right direction and that the model is not collapsing or degenerating.
The checkpoint we choose as the final model is not necessarily the one with the best performance with respect to the M-Prometheus-14B on the WMT test sets. We manually inspect the translation quality of the model on the WMT test sets and manually crafted adversarial examples to evaluate the robustness.

All the checkpoints of the 0.8B model show some weaknesses in handling certain challenging inputs. We apply linear merge combining three most recent checkpoints of MO-GRPO to mitigate it using Arcee's MergeKit \citep{goddard-etal-2024-arcees}.
The resulting merged model shows lower score on the WMT test sets but better on the adversarial examples, which we expect to be more useful for real-world applications.
For 7B model, we follow mostly the same procedure and hyperparameters as 3.3B model as we do not have sufficient computational resources to tune the training procedure for the 7B model. This may be the reason why it scores lower than the 3.3B model on the WMT test sets. Still, we find the model to generate higher quality text in manual inspection.

\section{Evaluation}
\label{sec:evaluation}

% We evaluate the performance of our translation models on a diverse set of benchmarks that cover various domains and text styles.
% \subsection{Benchmarks}
We evaluate our models on five translation benchmarks selected for two key criteria: (1) derivation from real-world applications rather than artificial test sets, and (2) less over-optimization in the community compared to widely-used datasets like WMT.
These are: (1) \textbf{Business Scene Dialogue (BSD)}~\citep{rikters-etal-2019-designing}: A corpus of business conversations originally collected for dialogue research. We translate each complete conversation rather than individual sentences to evaluate discourse-level translation. (2) \textbf{Court Interpreter (Court)}~\citep{yamagishi2025court}: Legal domain translations from Japanese court proceedings, testing formal register and domain terminology. (3) \textbf{JMedBench (JMed)}~\citep{lin2024post,jiang-etal-2025-jmedbench}: Medical domain translations, specifically the ejmmt (English-Japanese Medical Machine Translation) subsets, requiring specialized terminology. (4) \textbf{pfmt-bench-fin-ja (PFMT)}~\citep{hirano2025financial}: Financial domain translations covering business and economic content. (5) \textbf{WAT 2025 Patent Translation (PAT)}~\citep{nakazawa-etal-2025-findings}: Patent claim translations requiring technical accuracy and formal language.

% \begin{itemize}
%     \item \textbf{Business Scene Dialogue (BSD)}~\citep{rikters-etal-2019-designing}: A corpus of business conversations originally collected for dialogue research. We translate each complete conversation rather than individual sentences to evaluate discourse-level translation.
    
%     \item \textbf{Court Interpreter (Court)}~\citep{yamagishi2025court}: Legal domain translations from Japanese court proceedings, testing formal register and domain terminology. 
    
%     \item \textbf{JMedBench (JMed)}~\citep{jiang-etal-2025-jmedbench}: Medical domain translations, specifically the ejmmt (English-Japanese Medical Machine Translation) subsets, requiring specialized terminology.
    
%     \item \textbf{pfmt-bench-fin-ja (PFMT)}~\citep{hirano2025financial}: Financial domain translations covering business and economic content.
    
%     \item \textbf{WAT 2025 Patent Translation (PAT)}~\citep{nakazawa-etal-2025-findings}: Patent claim translations requiring technical accuracy and formal language.
% \end{itemize}

% \subsection{Results}

We evaluate the models with (1) BLEU scores computed using SacreBLEU~\citep{post-2018-call} and (2) M-Prometheus-14B.
The scores are compared against baseline models with strong translation performance on WMT benchmarks available at the moment of evaluation.
The results demonstrate that compact models can achieve competitive translation quality when trained with appropriate methodologies. 
For reproducibility, we use beam search decoding with a beam width of 1 by \texttt{transformers} library for all models except for plamo-2-translate.\footnote{plamo-2-translate uses the \texttt{vllm} library for decoding as our codebase has a compatibility issue with the \texttt{mamba-ssm} library.}
% Table~\ref{tab:results} shows the macro-average BLEU and M-Prometheus scores.
Tables~\ref{tab:bleu} and \ref{tab:prometheus} show the BLEU and Prometheus scores of the models on the benchmark tasks. The average shows the macro average.
Overall, our models show competitive performance, often outperforming much larger baseline models.

\paragraph{\bf Qualitative analysis.}
While the automatic metric scores provide a quantitative measure of translation quality, we also conduct a qualitative analysis to understand the strengths and weaknesses of our models in more depth. We manually inspect a sample of translations generated by our models across different benchmarks and compare them with the outputs from baseline models.
Table~\ref{tab:hy} shows a generation example by tencent/HY-MT1.5-7B on JMedBench. HY-MT1.5-7B is an upgraded version of the Hunyuan-MT-7B, which achieved the first place in 30 out of the 31 language categories it participated in at the WMT 2025~\citep{zheng2025hunyuanmttechnicalreport,zheng-etal-2025-shy,zheng2025hymttechnicalreport}. The output from HY-MT1.5-7B shows significant issues with repetition and failure to produce a coherent translation, while our 7B model generates a more accurate and fluent translation that closely follows the reference.
Because M-Prometheus scores the model from 1 to 5, generations with a few minor issues and detrimental issues can get the same score of 1, which makes it difficult to rely on its aggregated score for evaluating the performance of the translation models.
shisa-v2.1-llama3.2-3b also generates a good translation, but because the model is a general purpose model not specialized for translation, it tends to generate verbose explanation of the translation. This shows that the model has the potential to generate high-quality translations, but it is not production ready in a sense that it cannot be reliably used in real-world applications without further refinement. M-Prometheus gives it the score of 5, the same as our 7B model, which shows the limitation of relying on the aggregated score of M-Prometheus for evaluating translation quality in terms of practical usefulness.

\section{Conclusions}
\label{sec:conclusion}

The study presents an anecdotal demonstration that compact language models can achieve competitive Japanese-English translation quality for real-world applications using open source models and data. Through a carefully designed training pipeline that includes synthetic data generation, supervised fine-tuning, and reinforcement learning, we show that small models can outperform much larger baselines on diverse benchmarks derived from real-world scenarios.

% \clearpage

\section{Limitations}
\label{sec:limitations}

Our evaluation has several limitations. First, the benchmarks used for evaluation, while selected for their real-world relevance and lower likelihood of over-optimization, may still not fully capture the diversity of real-world translation scenarios. Second, the reliance on automatic metrics like BLEU and M-Prometheus, while practical under resource constraints, do not fully reflect human judgments of translation quality. For example, we find M-Prometheus to be less aligned for tasks outside of the WMT test sets and find BLEU to be more effective.
Third, the qualitative analysis is based on a limited number of examples and is subject to the authors' interpretations, which introduces bias.
Our models will be open source, and we encourage the community to conduct more comprehensive evaluations for a task of their interest.
We also advocate for the development of more real-world machine translation benchmarks curated by the practitioners in the respective domains and language pairs. 
Open sourcing more models in WMT submissions will also allow the community to stand on the shoulders of giants and conduct more comprehensive evaluations for a task of their interest.

The training procedure shows one successful instance of developing compact translation models.
We do not claim that the specific training pipeline we used is optimal or universally applicable. Alternative approaches to data synthesis, fine-tuning strategies, reward design, and base model selection may yield different results. Further research is needed to explore the design space of training methodologies for compact translation models.

Japanese is a resource-rich language compared to many other languages~\citep{joshi-etal-2020-state}. This allows us to train competitive translation models even with compact architectures. For lower-resource language pairs, the performance gap between small and large models may be wider, and the training methodologies may need to be adapted to account for data scarcity.

While the experiments are designed to run with limited computational resources, the training process still requires plenty of GPU-hours. This may limit the accessibility of this approach for researchers and practitioners working with limited computation or language resources.

\section*{Acknowledgment}

\begin{figure}
    \centering
    \includegraphics[width=0.7\linewidth, angle=-90]{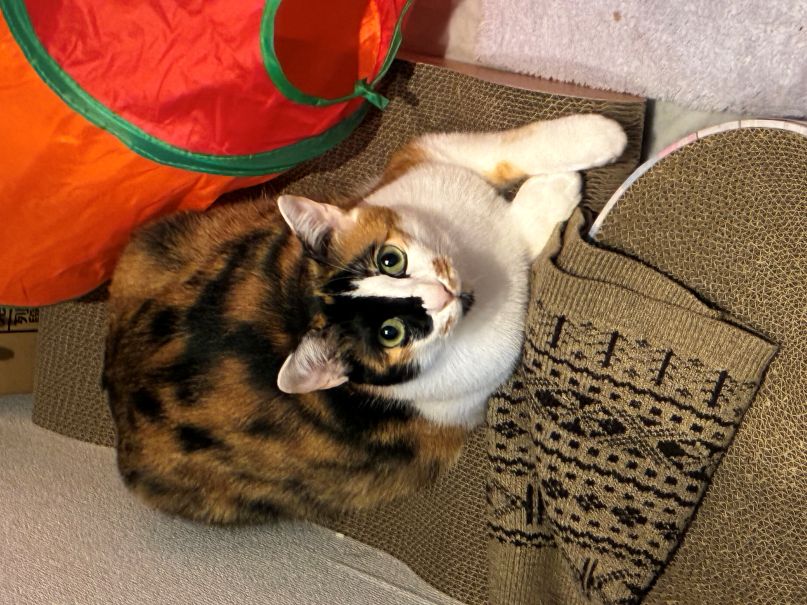}
    \caption{Queue, the cat, loves to sit on the keyboard and hit \emph{random} keys.}
    \label{fig:queue}
\end{figure}

We thank Mitsuki Sakamoto for deploying the model with UI for internal testing, which significantly help the manual inspection of the generations of the models.
Ryosuke Ishigami developed the base model of the 7B model and kindly shared it for this project.
We also thank the colleagues in CyberAgent AI Lab Reinforcement Learning Team for giving feedback to the project.
We thank Queue, the cat, for helping us understand the importance of backing up your manuscript frequently (Figure~\ref{fig:queue}).

\JPN{

% \bibliography{ms,anthology-1,anthology-2}
}

\appendix

\clearpage
\section{Hyperparameters}
\label{sec:appendix-hyperparams}

Tables~\ref{tab:hyperparams-sft1}, \ref{tab:hyperparams-sft2}, and \ref{tab:hyperparams-grpo} summarize the hyperparameters used for the first and second stages of SFT and MO-GRPO.
We do not perform extensive hyperparameter tuning due to limited computational resources, but we find that the selected hyperparameters work well for our training pipeline. The hyperparameters are shown for the 0.8B model; larger models use the similar hyperparameters except for the values that are adjusted for computational efficiency (e.g., batch size, gradient accumulation steps, max sequence length, etc.).

We use 8-bit Adam optimizer~\citep{or2025torchao} to reduce the VRAM consumption. In our preliminary evaluation, it shows negligible difference between a 32-bit Adam optimizer.
We find NEFTune~\citep{jain2024neftune} to be critical in stabilizing the SFT training in the preliminary experiments. We use it throughout the SFT stages.
Use of liger kernel~\citep{hsu2025ligerkernel} significantly contributed especially on training on document-level translation tasks by reducing the VRAM consumption of the loss computation.

\section{Prompt}
\label{appendix-prompt}
Although our models are specialized for machine translation, we employ an instruction-based format rather than direct source-to-target translation. This design choice provides better customizability, making it easier to extend the models for domain-specific applications or merge them with other instruction-tuned capabilities.
The input to the model follows the sarashina2.2-instruct chat template,\footnote{\url{https://huggingface.co/sbintuitions/sarashina2.2-1b-instruct-v0.1/blob/main/tokenizer_config.json\#L153}} and we use the following prompt for the translation:
\begin{verbatim}
Translate the following {src_lang} 
text into {tgt_lang}.

{src_text}
\end{verbatim}
where \texttt{src\_lang} and \texttt{tgt\_lang} are language names (``Japanese'' or ``English''), and \texttt{src\_text} is the input text to translate. The model generates the translation as a system response within the chat format.

We use this prompt for both training and evaluation to maintain consistency.

\begin{table}
\centering
\begin{tabular}{@{}lr@{}}
\toprule
Hyperparameter & Value \\
\midrule
Optimizer & 8 bit Adam \\
Learning rate & $2 \times 10^{-5}$ \\
LR scheduler & Linear \\
Warmup ratio & 0.015 \\
Max grad norm & 1.0 \\
Max sequence length & 768 \\
Per-device batch size & 32 \\
Gradient accumulation steps & 2 \\
Packing & No \\
NEFTune noise alpha & 5.0 \\
Liger Kernel & Yes \\
\bottomrule
\end{tabular}
\caption{Hyperparameters for the first stage of SFT on the 0.8B model.}
\label{tab:hyperparams-sft1}
\end{table}

\begin{table}
\centering
\begin{tabular}{@{}lr@{}}
\toprule
Hyperparameter & Value \\
\midrule
Optimizer & 8 bit Adam \\
Learning rate & $2 \times 10^{-5}$ \\
LR scheduler & Linear \\
Warmup ratio & 0.05 \\
Max grad norm & 1.0 \\
Max sequence length & 4,096 \\
Per-device batch size & 8 \\
Gradient accumulation steps & 8 \\
Packing & Yes \\
NEFTune noise alpha & 5.0 \\
Liger Kernel & Yes \\
\bottomrule
\end{tabular}
\caption{Hyperparameters for the second stage of SFT on the 0.8B model.}
\label{tab:hyperparams-sft2}
\end{table}

\begin{table}
\centering
\begin{tabular}{@{}lr@{}}
\toprule
Hyperparameter & Value \\
\midrule
Optimizer & 8 bit Adam \\
Learning rate & $1 \times 10^{-6}$ \\
LR scheduler & Linear \\
Warmup ratio & 0.01 \\
Per-device batch size & 16 \\
Gradient accumulation steps & 4 \\
Num generations per prompt & 16 \\
Max prompt length & 700 (4,096) \\
Max completion length & 700 (4,096) \\
Temperature & 1.0 \\
Top-$k$ & 80 \\
Repetition penalty & 1.05 \\
KL coefficient ($\beta$) & 0 \\
Liger Kernel & Yes \\
\bottomrule
\end{tabular}
\caption{Hyperparameters for MO-GRPO on the 0.8B model. Because 0.8B model has limited capability, we focus the model on improving the translation quality of shorter inputs and outputs, and therefore we set the max prompt length and max completion length to 700. For larger models, we set these values to 4,096 to allow the model to handle longer inputs and outputs.}
\label{tab:hyperparams-grpo}
\end{table}

\end{document}